\documentclass[10pt]{article}

\usepackage[letterpaper,textwidth=5.5in,textheight=9in,top=1in,
            headheight=12pt,headsep=25pt,footskip=30pt]{geometry}
\usepackage{times}
\makeatletter
\renewcommand{\normalsize}{%
  \@setfontsize\normalsize\@xpt\@xipt
  \abovedisplayskip      7\p@ \@plus 2\p@ \@minus 5\p@
  \abovedisplayshortskip \z@ \@plus 3\p@
  \belowdisplayskip      \abovedisplayskip
  \belowdisplayshortskip 4\p@ \@plus 3\p@ \@minus 3\p@}
\normalsize
\renewcommand{\small}{%
  \@setfontsize\small\@ixpt\@xpt
  \abovedisplayskip      6\p@ \@plus 1.5\p@ \@minus 4\p@
  \abovedisplayshortskip \z@ \@plus 2\p@
  \belowdisplayskip      \abovedisplayskip
  \belowdisplayshortskip 3\p@ \@plus 2\p@ \@minus 2\p@}
\renewcommand{\footnotesize}{\@setfontsize\footnotesize\@viiipt{9.5}}
\renewcommand{\scriptsize}{\@setfontsize\scriptsize\@viipt\@viiipt}
\renewcommand{\large}{\@setfontsize\large\@xiipt{14}}
\renewcommand{\Large}{\@setfontsize\Large\@xivpt{16}}
\renewcommand{\LARGE}{\@setfontsize\LARGE\@xviipt{20}}
\renewcommand{\section}{\@startsection{section}{1}{\z@}%
  {-2.0ex \@plus -0.5ex \@minus -0.2ex}{1.5ex \@plus 0.3ex \@minus 0.2ex}%
  {\large\bf\raggedright}}
\renewcommand{\subsection}{\@startsection{subsection}{2}{\z@}%
  {-1.8ex \@plus -0.5ex \@minus -0.2ex}{0.8ex \@plus 0.2ex}%
  {\normalsize\bf\raggedright}}
\renewcommand{\paragraph}{\@startsection{paragraph}{4}{\z@}%
  {1.5ex \@plus 0.5ex \@minus 0.2ex}{-1em}{\normalsize\bf}}
\renewcommand{\@maketitle}{%
  \vbox{%
    \hsize\textwidth \linewidth\hsize
    \vskip 0.1in
    \hrule height 4pt \vskip 0.25in \vskip -\parskip
    \centering {\LARGE\bf \@title\par}
    \vskip 0.29in \vskip -\parskip \hrule height 1pt \vskip 0.09in
    \def\And{\end{tabular}\hfil\linebreak[0]\hfil
             \begin{tabular}[t]{c}\bf\rule{\z@}{24\p@}\ignorespaces}%
    \begin{tabular}[t]{c}\bf\rule{\z@}{24\p@}\@author\end{tabular}%
    \vskip 0.3in \@minus 0.1in
  }%
}
\renewenvironment{abstract}{%
  \vskip 0.075in
  \centerline{\large\bf Abstract}
  \vspace{0.5ex}
  \begin{quote}}{\par\end{quote}\vskip 1ex}
\makeatother
\usepackage[utf8]{inputenc}
\usepackage[T1]{fontenc}
\usepackage[square]{natbib}
\setcitestyle{citesep={,}}
\usepackage[hidelinks]{hyperref}
\usepackage{url}
\usepackage{booktabs}
\usepackage{amsfonts}
\usepackage{amsmath}
\usepackage{nicefrac}
\usepackage{microtype}
\usepackage{xcolor}
\usepackage{graphicx}
\usepackage[font=small,skip=4pt]{caption}

\hypersetup{
  pdftitle={One-Step Prediction Fidelity Is a Poor Proxy for Planning Performance in JEPA World Models},
  pdfauthor={Joshua Cheung, Zhang He}
}

\newcommand{\Erepr}{E_{\mathrm{repr}}}
\newcommand{\Epred}{E_{\mathrm{pred}}}
\newcommand{\Eexcess}{E_{\mathrm{excess}}}
\newcommand{\enc}{\mathrm{enc}}
\newcommand{\pred}{\mathrm{pred}}

\title{One-Step Prediction Fidelity Is a Poor Proxy for\\
Planning Performance in JEPA World Models}

\author{Joshua Cheung \And Zhang He}

\begin{document}

\maketitle

\begin{abstract}
Latent world models are routinely selected, checkpointed and ablated on one-step
prediction loss, on the assumption that a model which predicts the next latent
better is a better model. In model-based RL this assumption is known to be
unreliable; we ask whether it also fails for \emph{latent} world models, where the
training target is the model's own embedding and the loss is not expressed in
interpretable units. We introduce a frozen-decoder protocol reporting one-step
error in physical units, decomposed into the part already present in the
encoder's representation and the part the predictor adds. Applying it to a recent
end-to-end JEPA world model across three continuous-control environments, we move
the metric by fine-tuning and measure goal-conditioned planning on the same
checkpoints, with three training seeds and paired planning seeds throughout.
Within one environment, a per-epoch ladder shows the relationship saturating
after a single epoch: the first epoch improves fidelity by $32\%$ and planning by
$4.4$\,pp, but across the fifteen checkpoints that follow --- a further $15\%$
fidelity improvement, monotone in every training seed --- the correlation between
one-step error and planning success is $r = -0.010$. This is not a task-ceiling
artifact: stratifying by a model-independent measure of training-data support,
planning in the lowest-support stratum starts at $80\%$, leaving ample headroom,
and is nonetheless flat from epoch~1 onward. Across environments the exchange rate
is equally unstable: a $93\%$ fidelity reduction on TwoRoom yields $-0.7$\,pp of
planning, while a $40\%$ reduction on PushT yields $+4.9$\,pp $[+1.1, +8.9]$. All
results replicate under a decoder-free measure, latent drift. We also show that a
matched continued-training control absorbs the entire fidelity gain we had
attributed to an auxiliary objective.
\end{abstract}

\section{Introduction}

Joint-Embedding Predictive Architectures (JEPAs) learn world models by predicting
the latent representation of a future observation rather than its pixels
\citep{lecun2022path}. Recent work has made end-to-end training from pixels stable
and cheap \citep{maes2026lewm}. Training is driven by a next-embedding prediction
loss, and that loss --- or a close proxy --- is also what practitioners use to
choose between checkpoints, to justify architectural changes, and to report
progress.

The implicit assumption is that lower one-step prediction error indicates a
better world model. In model-based reinforcement learning more broadly this
assumption is already known to fail: \citet{lambert2020objective} named the
\emph{objective mismatch} between model likelihood and control performance, and
showed that the likelihood of one-step-ahead predictions is not reliably
correlated with control performance, and the value-aware and value-equivalent
model-learning literature \citep{farahmand2017vaml,grimm2020value} is the
constructive response to it. That work concerns generative models trained by
likelihood or reconstruction. Whether the same mismatch afflicts \emph{latent}
world models --- where the training target is the model's own embedding rather
than an observation, and where the metric is therefore not even expressed in
interpretable units --- has not to our knowledge been measured.

There are reasons to expect it does. The prediction loss is computed in a latent
space of the model's own choosing, at a single step, under teacher forcing, on
trajectories drawn from the training distribution. Planning executes long action
sequences through the predictor, off-distribution, with no ground-truth latent to
anchor to at any point after the first.

Testing this requires three things that are individually unremarkable and rarely
combined. First, a measure of prediction fidelity that is interpretable and
comparable across models --- latent-space error norms are not, since each model
chooses its own latent scale. Second, an \emph{intervention} that moves that
measure, so the question becomes causal rather than correlational across an
uncontrolled model zoo. Third, a downstream measurement the intervention cannot
directly optimize, evaluated with enough statistical care to distinguish a real
effect from seed noise. Our contributions:

\begin{itemize}
  \item \textbf{A frozen-decoder protocol} (Section~\ref{sec:protocol}) reporting
  one-step error in ground-truth physical units. A single decoder, fitted only on
  encoder embeddings of real observations and then frozen, is applied both to real
  next-observation embeddings and to predicted next embeddings. Their difference
  isolates the predictor's contribution and detects when the protocol has no
  power.
  \item \textbf{A within-environment saturation result}
  (Section~\ref{sec:ladder}). A per-epoch checkpoint ladder on PushT shows
  fidelity improving monotonically across all three training seeds while planning
  goes flat after the first epoch ($r = -0.010$ over the fifteen post-epoch-1
  checkpoints). Stratifying by a model-independent measure of training-data
  support shows this is not a task-ceiling artifact, and identifies where these
  models fail: planning and one-step error are both substantially worse in regions
  where the offline data are sparse.
  \item \textbf{A cross-environment dissociation} (Section~\ref{sec:cross}), in
  which the largest fidelity improvement we induce buys the least planning.
  \item \textbf{A noise-floor calibration} (Section~\ref{sec:cautions}) showing
  untouched baselines varying by up to 26 percentage points across evaluation
  seeds --- larger than any effect we induce, and larger than differences
  routinely reported from single runs in this literature.
  \item \textbf{A methodological caution} (Section~\ref{sec:cautions}). An
  auxiliary state-grounding objective we introduced is indistinguishable from five
  additional epochs of ordinary training on every measure we computed. Reported
  without a matched control it would have appeared as a large, tightly-resolved
  improvement.
\end{itemize}

\section{Related work}

\paragraph{Objective mismatch.} \citet{lambert2020objective} showed that model
predictive accuracy and downstream control performance are only loosely coupled
in model-based RL, and that lower model loss can accompany worse control.
\citet{farahmand2017vaml} and \citet{grimm2020value} respond constructively by
training models against value-aware or value-equivalent objectives rather than
likelihood. Our contribution is not to observe the mismatch for the first time but
to measure it in the latent JEPA setting, where the training target is the model's
own embedding, with a controlled intervention and an interpretable metric.

\paragraph{JEPA world models.} \citet{lecun2022path} introduced the JEPA recipe;
self-supervised instantiations predict masked latent content for images
\citep{assran2023ijepa} and video \citep{assran2025vjepa2}, usually stabilized by
an EMA target encoder whose optimization semantics are unclear
\citep{ponce2026dual}. Action-conditioned variants either freeze a pretrained
encoder \citep{zhou2025dinowm} or train end-to-end with variance-covariance
regularization \citep{bardes2022vicreg,sobal2025pldm}. We evaluate LeWorldModel
\citep{maes2026lewm}: end-to-end from pixels with next-embedding prediction plus a
regularizer \citep{balestriero2025lejepa} enforcing approximately Gaussian latents
via random projections \citep{epps1983,cramer1936}.

\paragraph{Probing and physical structure.} A common evaluation trains probes to
recover physical quantities from learned representations
\citep{maes2026lewm,sobal2022slow}. A second family adopts the
violation-of-expectation paradigm \citep{margoni2024voe} to ask whether models
assign elevated surprise to physically implausible inputs
\citep{garrido2025intuitive,bordes2025intphys2}. Both measure what a
representation \emph{contains} or \emph{discriminates}, computed on encoder
outputs applied to real observations. We instead apply the decoder to the
predictor's output, measuring what the model \emph{produces}.

\paragraph{Planning with latent dynamics.} Model-based control either trains a
policy in imagination
\citep{ha2018worldmodels,hafner2020dreamer,hafner2025dreamerv3,hafner2025dreamer4}
or plans at test time with model-predictive control
\citep{hansen2022tdmpc,hansen2024tdmpc2,bar2025nwm}. We use the latter, with the
cross-entropy method \citep{rubinstein2004cem}.

\section{The frozen-decoder protocol}\label{sec:protocol}

\subsection{Setup and metrics}

Let $\enc_\theta$ map an observation $o_t$ to a latent $z_t$, and let
$\hat z_{t+1} = \pred_\phi(z_t, a_t)$. Let $s_t \in \mathbb{R}^k$ be the
ground-truth simulator state; angles are represented as $(\sin, \cos)$ pairs so
errors respect wrap-around.

We fit a decoder $D$ from latents to physical state using \emph{only} encoder
embeddings of \emph{real} observations, on a held-out set of training episodes,
with the world model frozen. $D$ is then frozen and applied to two different
inputs on disjoint test episodes:
\begin{equation}
  \Erepr = d\big(D(\enc_\theta(o_{t+1})),\, s_{t+1}\big), \qquad
  \Epred = d\big(D(\hat z_{t+1}),\, s_{t+1}\big), \qquad
  \Eexcess = \Epred - \Erepr,
  \label{eq:metrics}
\end{equation}
where $d$ is squared error per state component, divided by that component's
training-set variance and averaged. Normalisation statistics are shared by all
models within an environment.

$\Erepr$ asks how readable the next state is from the encoder's representation of
the real next observation; it involves the predictor not at all. $\Epred$ asks the
same of the predictor's output given the true action, and is the quantity a
practitioner has in mind when speaking of one-step prediction quality. $\Eexcess$
is the one we argue should be reported: if the predictor were perfect,
$\hat z_{t+1} = z_{t+1}$ and $\Eexcess = 0$ exactly, whatever $D$ does.
Conversely, when $\Erepr \approx \Epred$, almost all measured error originates
before the predictor is involved and the protocol has little power --- a
condition we observe on one environment (Section~\ref{sec:nondegenerate}) and
which $\Epred$ alone would have concealed.

\subsection{Is the decoder extrapolating?}\label{sec:extrapolation}

$D$ is fitted on $\enc_\theta(o)$ and evaluated on $\hat z = \pred_\phi(z, a)$. If
predicted latents sit off the encoder's output manifold, $D$ is being asked to
extrapolate and $\Eexcess$ could partly measure manifold offset rather than
physical inaccuracy.

We check this against latent drift $\|\hat z_{t+1} - z_{t+1}\|_2$, which never
passes through the decoder. Within a single environment --- across the fifteen
PushT ladder checkpoints of Section~\ref{sec:ladder}, where scales are comparable
--- mean $\Eexcess$ and mean latent drift correlate at $r = 0.85$. The two
therefore track each other at the level our claims are made, and every result
below replicates under a decoder-free measure. (The same correlation computed
across all twelve environment--variant combinations is $0.99$, but those points
span orders of magnitude in both quantities, so that figure is largely driven by
between-environment scale and we do not rely on it.) Per transition the two are
essentially uncorrelated (PushT $r = 0.012$, OGBench-Cube $r = -0.007$; TwoRoom
is the exception at $r = 0.34$--$0.59$, plausibly because its state is
two-dimensional and drift is dominated by the position axis), so they are not
mechanically coupled within a model. Drift also exhibits the same dissociation
independently: it falls $77\%$ on TwoRoom for $-0.7$\,pp of planning and $32\%$
on PushT for $+4.9$\,pp.

\subsection{Interventions and planning evaluation}\label{sec:interventions}

We warm-start from released checkpoints and continue training for 5 epochs at
learning rate $10^{-5}$, batch size 128, with three training seeds per
environment. \textbf{Continued} uses the original two-term objective only.
\textbf{State-grounded} adds an auxiliary head predicting standardized physical
state from encoder embeddings (weight $0.1$, discarded after training); it was
initially proposed as an improvement and is reported here as an instrument.
Section~\ref{sec:cautions} shows the two are not distinguishable, so the induced
variation should be understood as \emph{one} intervention under two labels ---
five additional epochs --- rather than two independent manipulations.

\begin{figure}[t]
  \centering
  \includegraphics[width=0.62\textwidth]{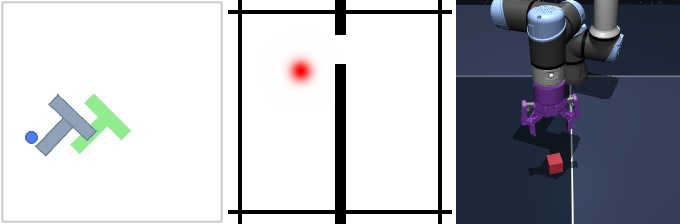}
  \caption{Evaluation environments. Left to right: PushT, TwoRoom, OGBench-Cube.}
  \label{fig:envs}
\end{figure}

A further variant adding auxiliary supervision on state \emph{differences} was
evaluated on fidelity but not planning, and is excluded from the analysis; it
changes no conclusion. Because the auxiliary head and the evaluation decoder share
a target parameterization, an improvement in $\Erepr$ under grounding is close to
guaranteed by construction; we treat it as a manipulation check, not a result.
The auxiliary loss is applied to encoder embeddings only, so the predictor is
never directly supervised toward decodability.

Planning uses the released evaluation unchanged: cross-entropy method
\citep{rubinstein2004cem} under receding-horizon MPC over 50 goal-conditioned test
trajectories. Solver, hyperparameters and test trajectories are fixed across
variants, and the same three planning seeds are used at every training seed, so
all differences are paired.

\section{Experimental setup}\label{sec:setup}

\paragraph{Environments.} PushT (2D manipulation, 5 state dimensions, $41{,}983$
test transitions); TwoRoom (2D navigation through a doorway, 2 dimensions,
$16{,}076$); OGBench-Cube \citep{park2025ogbench} (3D manipulation, 9 dimensions,
$38{,}000$). Figure~\ref{fig:envs} shows representative frames. All are
continuous-action and trained offline without reward.

\paragraph{Decoder.} A two-layer MLP, hidden width 512, up to 60 epochs with early
stopping (patience 10), batch size $4096$. No test episode contributes to fitting
$D$.

\paragraph{Statistics.} Three training seeds per variant per environment, five
paired decoder/data-split seeds for fidelity, three paired planning seeds. Point
estimates average transitions within episodes, then episodes, then evaluation
seeds, then training seeds. Intervals come from $10{,}000$ hierarchical paired
bootstrap repetitions resampling at all three levels while preserving variant
pairing.

\paragraph{Limits of that procedure.} Two limitations should be stated up front.
A bootstrap resampling only three training seeds gives poorly calibrated
intervals at the top level, and our planning intervals should be read as
indicative rather than exact. And planning success is a binomial over 50
trajectories, which places a floor of roughly 4--7\,pp on what any single
evaluation can resolve. Effects smaller than that are not measurable in our
design at any seed count, and we do not claim to measure them.

\section{Are the predictors non-degenerate?}\label{sec:nondegenerate}

Replacing the true action with one sampled from a different transition raises
$\Epred$ by $29.1\times$ on PushT, $15.7\times$ on TwoRoom and $2.1\times$ on
OGBench-Cube: the predictors are genuinely action-conditioned. Against trivial
baselines --- persistence ($\hat s_{t+1} = s_t$) and constant velocity
($\hat s_{t+1} = s_t + (s_t - s_{t-1})$), both computed on ground-truth state ---
PushT and TwoRoom exceed skill $0.87$.

OGBench-Cube is the exception. Its constant-velocity skill is $-0.74$: the learned
predictor is \emph{worse} than linear extrapolation. Its action sensitivity is
weakest, and its $\Eexcess$ is only $0.3\%$ of $\Epred$, meaning almost all
measured error originates in the decoder. The protocol has little power there,
and $\Eexcess$ detects this without reference to any downstream task. We include
OGBench-Cube below but weight it accordingly.

\begin{table}[t]
  \centering
  \small
  \begin{tabular}{rrrrrrrr}
    \toprule
    & & & & \multicolumn{4}{c}{Planning success (\%)} \\
    \cmidrule(l){5-8}
    Epoch & $\Epred$ & $\Delta$ (\%) & Drift & Low sup. & Med.\ sup. & High sup. & All \\
    \midrule
    0 & 0.003336 & ---     & 1.123 & 80.00 & 90.14 & 100.00 & 90.00 \\
    1 & 0.002268 & $-32.0$ & 0.839 & 89.17 & 94.84 &  99.15 & 94.44 \\
    2 & 0.002109 & $-36.8$ & 0.814 & 85.83 & 95.78 &  97.44 & 93.56 \\
    3 & 0.002071 & $-37.9$ & 0.787 & 86.67 & 95.78 &  99.15 & 94.22 \\
    4 & 0.002026 & $-39.3$ & 0.774 & 88.33 & 95.78 &  99.15 & 94.67 \\
    5 & 0.002013 & $-39.7$ & 0.768 & 87.50 & 96.71 &  98.29 & 94.67 \\
    \bottomrule
  \end{tabular}
  \caption{PushT per-epoch ladder, three training seeds. Fidelity and latent drift
  fall monotonically throughout. Planning is flat after epoch~1 in every support
  stratum. Support groups are defined in Section~\ref{sec:ceiling} from the initial
  block position only, independently of any model. Baseline units per stratum: 40
  low, 71 medium, 39 high; a single trajectory--seed unit is 0.5--0.9\,pp.}
  \label{tab:ladder}
\end{table}

\section{Within one environment: the relationship saturates}\label{sec:ladder}

Our fine-tuning runs save a checkpoint each epoch, giving a graded fidelity ladder
inside a single environment at evaluation cost only. Table~\ref{tab:ladder}
reports PushT, averaged over three training seeds, with five decoder seeds and
three paired planning seeds per checkpoint.

The first epoch moves both quantities: $-32.0\%$ fidelity, $+4.44$\,pp planning.
Read alone this is consistent with the proxy behaving as assumed, and we report
it as such.

The remaining four epochs are the informative part. Fidelity falls a further
$11.2\%$ and latent drift a further $8.5\%$, both monotonically and on every
individual training seed. Aggregate planning moves $+0.22$\,pp, and the per-seed
changes from epoch~1 to epoch~5 are $+2.00$, $0.00$ and $-1.33$, differing in
sign. Because a $0.22$\,pp shift is far below what this evaluation can resolve
(Section~\ref{sec:setup}), we do not rest the claim on that magnitude. The claim
rests on the correlation over the graded portion of the ladder. Across the
fifteen post-epoch-1 checkpoints --- epochs 1--5 at three training seeds, spanning
a further $15\%$ range of $\Epred$ within a single task --- the correlation
between one-step error and planning success is $r = -0.010$, and between latent
drift and planning success $r = -0.175$.

Including the baseline checkpoint raises this to $r = -0.88$ over the full $40\%$
range. We report that figure but do not rest on it: it is carried entirely by the
single epoch $0 \to 1$ step, and because the baseline was evaluated once and
shared across training seeds, the three added points are identical and act as
one lever. The informative comparison is between that first step, which moves
both quantities, and everything after it, which moves only one.

\subsection{This is not a task ceiling}\label{sec:ceiling}

Aggregate PushT planning is $90\%$ at epoch~0, so an obvious objection is that the
task is saturated rather than the relationship. We test this with a difficulty
measure that is defined without reference to any model.

Following an analysis of the offline dataset, we partition the $512 \times 512$
workspace into an $8 \times 8$ grid and count the dataset states whose block
centre falls in each cell. Weighted quartiles of these counts define low, medium
and high support strata, covering $25.2\%$, $48.5\%$ and $26.3\%$ of dataset
states. The partition is spatial: low-support states are edge-like (mean
$160.7$\,px from the workspace centre) and high-support states centre-like
($38.7$\,px). Test trajectories are assigned by the block position at their
initial state.

The pretrained baseline is strongly stratified by this measure. Under a
ten-planning-seed evaluation, high-support planning exceeds low-support planning
by $18.28$\,pp $[+2.38, +39.35]$, and the high-minus-low contrast in one-step
block-position error is $-0.002127$ $[-0.002353, -0.001927]$; the three-seed
ladder of Table~\ref{tab:ladder} gives a $20.0$\,pp planning gap at epoch~0.
\emph{Where the training data are sparse, the model both predicts worse and plans
worse} --- a positive finding about where these models fail, independent of
anything else here.

It also supplies the headroom the ceiling objection requires. Low-support
planning begins at $80.0\%$, some 20\,pp below saturation. If the aggregate
flatness were a ceiling effect, this is the stratum in which fidelity gains should
convert into planning gains. They do not: low-support planning peaks at epoch~1
($+9.17$\,pp), then declines and never recovers, ending at $+7.50$\,pp while
fidelity continues to improve monotonically. The epoch-1-to-5 change of
$-1.67$\,pp is two trajectories out of 120 comparisons and we do not interpret its
sign; what matters is that it is not an increase. Medium support behaves the same
way, flat from epoch~2. High support, where a genuine ceiling does operate, is
negative throughout ($-0.85$ to $-2.56$\,pp from a baseline of $100\%$).

Per-trajectory outcomes confirm that the flat aggregate is not cancelling churn of
growing magnitude. Relative to epoch~0, trajectories that flip from success to
failure number 8, 10, 7, 7, 10 across epochs 1--5 out of 405 baseline successes
--- real backward movement, roughly $2\%$, with no trend as fidelity improves.

\begin{table}[t]
  \centering
  \footnotesize
  \setlength{\tabcolsep}{4pt}
  \begin{tabular}{llrrr}
    \toprule
    Env & Intervention & $\Delta\Epred$ (\%) & $\Delta\Eexcess$ (\%) & $\Delta$ planning (pp) \\
    \midrule
    TwoRoom & Continued      & $-93.1$ $[-93.8, -92.3]$ & $-93.4$ $[-94.1, -92.6]$ & $-0.67$ $[-4.22, +3.12]$ \\
    TwoRoom & State-grounded & $-91.8$ $[-93.0, -90.1]$ & $-92.1$ $[-93.3, -90.4]$ & $-0.44$ $[-3.56, +2.89]$ \\
    PushT   & State-grounded & $-42.2$ $[-54.6, -28.3]$ & $-57.0$ $[-68.2, -48.2]$ & $+3.56$ $[-0.22, +7.56]$ \\
    PushT   & Continued      & $-39.7$ $[-44.8, -33.8]$ & $-61.5$ $[-74.2, -52.3]$ & $\mathbf{+4.89}$ $\mathbf{[+1.11, +8.89]}$ \\
    Cube    & State-grounded & $-1.0$ $[-1.5, -0.6]$    & $-54.8$ $[-72.7, -34.3]$ & $+2.00$ $[-2.22, +6.01]$ \\
    Cube    & Continued      & $-0.3$ $[-0.6, +0.0]$    & $-90.4$ $[-110.1, -70.3]^\dagger$ & $+1.11$ $[-4.67, +6.67]$ \\
    \bottomrule
  \end{tabular}
  \caption{Change relative to baseline, $95\%$ hierarchical paired-bootstrap
  intervals over three training seeds, ordered by fidelity improvement. Only the
  bolded planning interval excludes zero. Between-arm differences quoted in
  Section~\ref{sec:cautions} use unrounded values and may differ in the last digit
  from subtracting the entries shown. $^\dagger$Cube's $\Eexcess$ is $0.0007$ at
  baseline and $0.0002$ after continued training, against $\Erepr = 0.244$;
  resampled estimates go negative when decoder noise exceeds the predictor's
  contribution, so the interval extends past $-100\%$. See
  Section~\ref{sec:nondegenerate}.}
  \label{tab:cross}
\end{table}

\subsection{A third intervention: rebalancing data coverage}\label{sec:rebalance}

The support stratification suggests a further intervention that manipulates
\emph{data} rather than the objective. Holding checkpoint, architecture,
objective, sample count and epochs fixed, we resampled the $200{,}000$ selected
training windows to favour rare block positions, raising the low-support share
from $27.1\%$ to $43.6\%$.

Its pre-specified outcomes were null: low-support block-position error changed by
$-0.000036$ $[-0.000101, +0.000032]$ and low-support planning by $+1.245$\,pp
$[-3.611, +5.989]$. A secondary, uncorrected outcome did move --- low-support
mean state-prediction nMSE fell by $0.000207$ $[-0.000386, -0.000019]$, a $6.7\%$
relative reduction --- so rebalancing changes some aspects of one-step physical
prediction without producing detectable planning improvement. We report it as a
third instrument exhibiting the same signature, not as a positive result.

\section{Across environments: no stable exchange rate}\label{sec:cross}

Table~\ref{tab:cross} orders the six intervention--environment pairs by fidelity
improvement. The two largest, both above $91\%$, produce the only two
\emph{negative} planning estimates. The single resolvable planning improvement,
$+4.9$\,pp, arises from a fidelity change less than half that size --- and belongs
to the control arm, not the intervention. OGBench-Cube's $-0.3\%$ produces a
larger planning point estimate than TwoRoom's $-93.1\%$.

\paragraph{We are explicit about what this comparison is.} Because the two
interventions are indistinguishable (Section~\ref{sec:cautions}), the fidelity
variation in Table~\ref{tab:cross} is almost entirely \emph{between}
environments: within each environment the two rows differ by a few percentage
points of fidelity and produce same-signed planning changes. Neither side of this
comparison is strictly commensurable across tasks. Planning success differs in
baseline and headroom; equally, a percentage reduction in $\Epred$ does not mean
the same thing in a two-dimensional state space whose $\Eexcess$ is essentially
all of $\Epred$ as it does in a nine-dimensional one whose predictor is
degenerate. Table~\ref{tab:cross} alone would therefore support only a weak
claim. It is Section~\ref{sec:ladder}, where fidelity varies within a single
task, that supplies the within-environment evidence; the two together establish
that the exchange rate between fidelity and planning is neither stable across
tasks nor sustained within one.

\begin{figure}[t]
  \centering
  \includegraphics[width=0.62\textwidth]{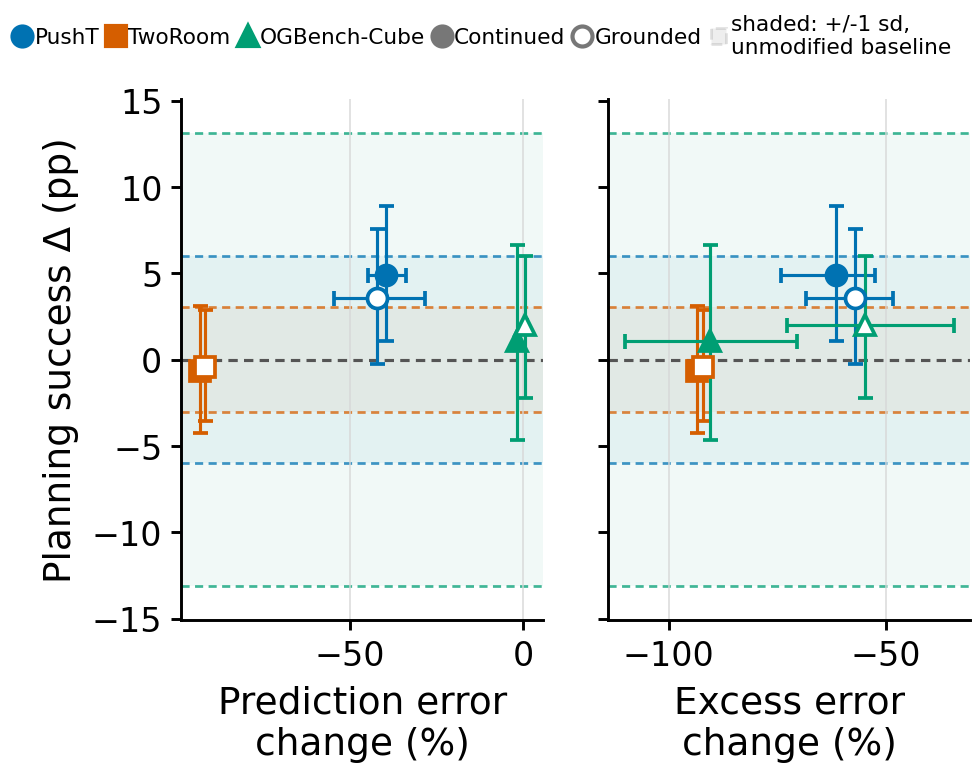}
  \caption{Change in one-step fidelity (left: $\Epred$; right: $\Eexcess$) against
  paired change in planning success, three training seeds. Shaded bands show
  $\pm 1$ standard deviation of the \emph{unmodified} baseline across evaluation
  seeds, one band per environment (widest: OGBench-Cube, $\pm 13.1$). Five of six
  points lie within the ambient variation of doing nothing.}
  \label{fig:tradeoff}
\end{figure}

Figure~\ref{fig:tradeoff} plots both fidelity measures against paired planning
change.

\section{Two ways to see an effect that is not there}\label{sec:cautions}

Two features of this design would, if omitted, have produced confident
conclusions from nothing. We report them separately from the main result because
they stand independently of it.

\textbf{The planning noise floor.} Evaluating the untouched baseline checkpoints
across planning seeds --- changing nothing about the model --- gives 84--96 on
PushT (sd $6.0$), 82--88 on TwoRoom (sd $3.1$) and 58--84 on OGBench-Cube (sd
$13.1$). The unmodified OGBench-Cube baseline varies by 26 percentage points
across evaluation seeds alone, against a largest observed intervention effect
anywhere of $4.9$\,pp. Planning comparisons in this literature are frequently
reported from single runs. At these variances a single-seed comparison can
produce an apparent effect of ten or more percentage points from nothing, in
either direction: our own OGBench-Cube grounded-minus-baseline differences are
$+6$, $-4$ and $+6$ across three planning seeds at one training seed. Pairing
across matched seeds is what recovers enough power to resolve the one effect we
can resolve. Paired multi-seed evaluation should be the default for this class of
comparison, and the seed spread should be reported alongside the mean.

\textbf{A matched control absorbs the intervention.} State grounding is not
distinguishable from continued training on any measure we computed. On TwoRoom,
Continued alone achieves $-93.1\%$ against grounding's $-91.8\%$ --- the control
is nominally \emph{better}, with overlapping intervals. Paired planning
differences between the arms, resampled over training seeds, are $-1.33$
$[-4.22, +1.78]$ on PushT, $+0.22$ $[-4.44, +4.00]$ on TwoRoom and $+0.89$
$[-4.00, +5.11]$ on OGBench-Cube: no consistent sign, all containing zero. Latent
drift does not separate them either. We flag this because the control is easy to
omit, and its omission is not obviously negligent. The two arms differ in one loss
weight; the auxiliary objective is well-motivated; the improvement is large,
tightly resolved, and reproduces across five evaluation and three training seeds.
Reported without a matched continued-training arm, the natural conclusion would
be that an auxiliary state-grounding loss reduces one-step physical prediction
error by more than $90\%$. That number is real. Its attribution would have been
wrong.

\section{Discussion}\label{sec:discussion}

\textbf{What we are entitled to conclude.} We do not claim that fidelity
interventions never affect planning: one of our six effects is resolved and
positive, and the first epoch of fine-tuning moves both quantities together. What
we can exclude is a stable, monotone exchange rate of the kind the proxy's use
presupposes. Within PushT the relationship saturates after one epoch; across
environments, a $93\%$ fidelity reduction on TwoRoom buys less planning than a
$0.3\%$ reduction on OGBench-Cube does. Neither pattern licenses ranking
checkpoints or architectures by one-step loss.

\textbf{Why this might be.} We advance no mechanism, having measured none.
One-step teacher-forced error may be dominated by transitions planning never
visits; planning may be limited by the optimizer or by compounding error rather
than single-step accuracy; or the planning cost may be insensitive to the latent
directions our interventions improve. Distinguishing these requires rollout-level
measurement, to which the protocol extends directly.

\textbf{Limitations.} Our fidelity measure is one-step and teacher-forced while
planning is long-horizon and off-distribution; a rollout-level version of this
measurement is the experiment that would settle how far the result generalizes,
and we regard it as the most promising direction. We evaluate one architecture at
15M parameters, three low-dimensional simulated environments, one planner, and
warm-started fine-tunes from one checkpoint family, so we characterize a
neighbourhood rather than the space of world models. Our within-environment
ladder covers one environment; the correlation is estimated over the $15\%$
fidelity range lying beyond the first epoch, which is about a fifth of the total
baseline-to-epoch-5 reduction, and a wider within-task sweep via auxiliary-weight
or capacity ablation would strengthen it. Our support stratification addresses the
task-ceiling confound but rests on 40 low-support baseline units, so individual
stratum differences of one or two percentage points are not interpretable. Three
training seeds and 50 planning trajectories bound what we can resolve, as noted in
Section~\ref{sec:setup}.

\textbf{Recommendations.} Do not present a reduction in latent prediction loss as
evidence of a better world model without a downstream measurement. When a
fidelity metric is reported, decompose it: $\Eexcess$ separates predictor error
from decoder error and identifies when a probing-based evaluation has no power.
Accompany any fine-tuning intervention with a matched continued-training arm.
Report planning with paired multiple seeds at both the training and evaluation
levels, and publish the seed spread.

\section{Conclusion}

We introduced a frozen-decoder protocol measuring one-step world-model prediction
error in physical units and separating the predictor's contribution from the
encoder's. Moving that metric by fine-tuning, we found the relationship to
goal-conditioned planning saturates within a task after a coarse initial gain,
and has no stable exchange rate across tasks: the largest fidelity improvements we
induced bought the least planning. The result replicates under a decoder-free
measure, and a matched continued-training control absorbs the entire improvement
we had attributed to an auxiliary objective. One-step fidelity is a poor proxy for
planning competence at this scale.

{\small
\bibliographystyle{abbrvnat}
\bibliography{references}

@inproceedings{assran2023ijepa,
  author    = {Assran, Mahmoud and Duval, Quentin and Misra, Ishan and Bojanowski, Piotr and Vincent, Pascal and Rabbat, Michael and LeCun, Yann and Ballas, Nicolas},
  title     = {Self-supervised learning from images with a joint-embedding predictive architecture},
  booktitle = {CVPR},
  pages     = {15619--15629},
  year      = {2023}
}

@article{assran2025vjepa2,
  author  = {Assran, Mido and Bardes, Adrien and Fan, David and Garrido, Quentin and Howes, Russell and Muckley, Matthew and Rizvi, Ammar and Roberts, Claire and Sinha, Koustuv and Zholus, Artem and others},
  title   = {{V-JEPA} 2: Self-supervised video models enable understanding, prediction and planning},
  journal = {arXiv:2506.09985},
  year    = {2025}
}

@article{balestriero2025lejepa,
  author  = {Balestriero, Randall and LeCun, Yann},
  title   = {{LeJEPA}: Provable and scalable self-supervised learning without the heuristics},
  journal = {arXiv:2511.08544},
  year    = {2025}
}

@inproceedings{bar2025nwm,
  author    = {Bar, Amir and Zhou, Gaoyue and Tran, Danny and Darrell, Trevor and LeCun, Yann},
  title     = {Navigation world models},
  booktitle = {CVPR},
  pages     = {15791--15801},
  year      = {2025}
}

@inproceedings{bardes2022vicreg,
  author    = {Bardes, Adrien and Ponce, Jean and LeCun, Yann},
  title     = {{VICReg}: Variance-invariance-covariance regularization for self-supervised learning},
  booktitle = {ICLR},
  year      = {2022}
}

@article{bordes2025intphys2,
  author  = {Bordes, Florian and Garrido, Quentin and Kao, Justine T. and Williams, Adina and Rabbat, Michael and Dupoux, Emmanuel},
  title   = {{IntPhys} 2: Benchmarking intuitive physics understanding in complex synthetic environments},
  journal = {arXiv:2506.09849},
  year    = {2025}
}

@article{cramer1936,
  author  = {Cram{\'e}r, Harald and Wold, Herman},
  title   = {Some theorems on distribution functions},
  journal = {Journal of the London Mathematical Society},
  volume  = {1},
  number  = {4},
  pages   = {290--294},
  year    = {1936}
}

@article{epps1983,
  author  = {Epps, T. W. and Pulley, L. B.},
  title   = {A test for normality based on the empirical characteristic function},
  journal = {Biometrika},
  volume  = {70},
  number  = {3},
  pages   = {723--726},
  year    = {1983}
}

@inproceedings{farahmand2017vaml,
  author    = {Farahmand, A. and Barreto, Andr{\'e} and Nikovski, Daniel},
  title     = {Value-aware loss function for model-based reinforcement learning},
  booktitle = {Artificial Intelligence and Statistics (AISTATS)},
  pages     = {1486--1494},
  publisher = {PMLR},
  year      = {2017}
}

@article{garrido2025intuitive,
  author  = {Garrido, Quentin and Ballas, Nicolas and Assran, Mahmoud and Bardes, Adrien and Najman, Laurent and Rabbat, Michael and Dupoux, Emmanuel and LeCun, Yann},
  title   = {Intuitive physics understanding emerges from self-supervised pretraining on natural videos},
  journal = {arXiv:2502.11831},
  year    = {2025}
}

@inproceedings{grimm2020value,
  author    = {Grimm, Christopher and Barreto, Andr{\'e} and Singh, Satinder and Silver, David},
  title     = {The value equivalence principle for model-based reinforcement learning},
  booktitle = {Advances in Neural Information Processing Systems},
  volume    = {33},
  pages     = {5541--5552},
  year      = {2020}
}

@article{ha2018worldmodels,
  author  = {Ha, David and Schmidhuber, J{\"u}rgen},
  title   = {World models},
  journal = {arXiv:1803.10122},
  year    = {2018}
}

@inproceedings{hafner2020dreamer,
  author    = {Hafner, Danijar and Lillicrap, Timothy and Ba, Jimmy and Norouzi, Mohammad},
  title     = {Dream to control: Learning behaviors by latent imagination},
  booktitle = {ICLR},
  year      = {2020}
}

@article{hafner2025dreamerv3,
  author  = {Hafner, Danijar and Pasukonis, Jurgis and Ba, Jimmy and Lillicrap, Timothy},
  title   = {Mastering diverse control tasks through world models},
  journal = {Nature},
  volume  = {640},
  number  = {8059},
  pages   = {647--653},
  year    = {2025}
}

@article{hafner2025dreamer4,
  author  = {Hafner, Danijar and Yan, Wilson and Lillicrap, Timothy},
  title   = {Training agents inside of scalable world models},
  journal = {arXiv:2509.24527},
  year    = {2025}
}

@inproceedings{hansen2022tdmpc,
  author    = {Hansen, Nicklas and Wang, Xiaolong and Su, Hao},
  title     = {Temporal difference learning for model predictive control},
  booktitle = {ICML},
  year      = {2022}
}

@inproceedings{hansen2024tdmpc2,
  author    = {Hansen, Nicklas and Su, Hao and Wang, Xiaolong},
  title     = {{TD-MPC2}: Scalable, robust world models for continuous control},
  booktitle = {ICLR},
  year      = {2024}
}

@inproceedings{lambert2020objective,
  author    = {Lambert, Nathan and Amos, Brandon and Yadan, Omry and Calandra, Roberto},
  title     = {Objective mismatch in model-based reinforcement learning},
  booktitle = {Learning for Dynamics and Control (L4DC)},
  series    = {PMLR},
  volume    = {120},
  pages     = {761--770},
  year      = {2020}
}

@misc{lecun2022path,
  author       = {LeCun, Yann},
  title        = {A path towards autonomous machine intelligence, version 0.9.2},
  howpublished = {OpenReview},
  year         = {2022}
}

@article{maes2026lewm,
  author  = {Maes, Lucas and {Le Lidec}, Quentin and Scieur, Damien and LeCun, Yann and Balestriero, Randall},
  title   = {{LeWorldModel}: Stable end-to-end joint-embedding predictive architecture from pixels},
  journal = {arXiv:2603.19312},
  year    = {2026}
}

@article{margoni2024voe,
  author  = {Margoni, Francesco and Surian, Luca and Baillargeon, Ren{\'e}e},
  title   = {The violation-of-expectation paradigm: A conceptual overview},
  journal = {Psychological Review},
  volume  = {131},
  number  = {3},
  pages   = {716},
  year    = {2024}
}

@inproceedings{park2025ogbench,
  author    = {Park, Seohong and Frans, Kevin and Eysenbach, Benjamin and Levine, Sergey},
  title     = {{OGBench}: Benchmarking offline goal-conditioned {RL}},
  booktitle = {ICLR},
  year      = {2025}
}

@inproceedings{ponce2026dual,
  author    = {Ponce, Jean and Terver, Basile and Hebert, Martial and Arbel, Michael},
  title     = {Dual perspectives on non-contrastive self-supervised learning},
  booktitle = {ICLR},
  year      = {2026}
}

@book{rubinstein2004cem,
  author    = {Rubinstein, Reuven Y. and Kroese, Dirk P.},
  title     = {The cross-entropy method: a unified approach to combinatorial optimization, {Monte-Carlo} simulation and machine learning},
  publisher = {Springer},
  year      = {2004}
}

@article{sobal2022slow,
  author  = {Sobal, Vlad and {Jyothir S V} and Jalagam, Siddhartha and Carion, Nicolas and Cho, Kyunghyun and LeCun, Yann},
  title   = {Joint embedding predictive architectures focus on slow features},
  journal = {arXiv:2211.10831},
  year    = {2022}
}

@inproceedings{sobal2025pldm,
  author    = {Sobal, Vlad and Zhang, Wancong and Cho, Kyunghyun and Balestriero, Randall and Rudner, Tim G. J. and LeCun, Yann},
  title     = {Learning from reward-free offline data: A case for planning with latent dynamics models},
  booktitle = {Advances in Neural Information Processing Systems},
  volume    = {38},
  year      = {2025}
}

@inproceedings{zhou2025dinowm,
  author    = {Zhou, Gaoyue and Pan, Hengkai and LeCun, Yann and Pinto, Lerrel},
  title     = {{DINO-WM}: World models on pre-trained visual features enable zero-shot planning},
  booktitle = {ICML},
  pages     = {79115--79135},
  publisher = {PMLR},
  year      = {2025}
}
}

\appendix

\section{Absolute values and reproducibility notes}\label{app:absolute}

All claims in the main text are relative changes with intervals. For reference,
baseline (epoch-0) values, averaged over five evaluation seeds at the reference
training seed, are: PushT $\Erepr = 0.00260$, $\Epred = 0.00334$, drift $1.123$;
TwoRoom $2.99 \times 10^{-5}$, $0.01017$, $3.439$; OGBench-Cube $0.24421$,
$0.24492$, $0.862$. Persistence and constant-velocity skill are $0.945/0.944$
(PushT), $0.871/0.914$ (TwoRoom) and $-0.637/-0.736$ (OGBench-Cube).

Hyperparameters match across environments and seeds, verified by auditing every
saved run configuration. One state-grounded checkpoint needed a parameter-name
remap for an older ViT key layout, verified lossless (303/303 tensors, zero value
mismatches). Baseline PushT planning matches the originally reported figure on one
of three evaluation seeds ($96\%$), averaging $90\%$. The ladder used only
checkpoints saved during the original runs; no extra training was done.

\end{document}